\documentclass[runningheads,a4paper]{llncs}

\usepackage{amssymb}
\usepackage{graphicx}

\usepackage{verbatim}
\usepackage{booktabs}
\usepackage{multirow}
\usepackage{amsmath}

\usepackage{url}
\urldef{\mailsa}\path|{alfred.hofmann, ursula.barth, ingrid.haas, frank.holzwarth,|
\urldef{\mailsb}\path|anna.kramer, leonie.kunz, christine.reiss, nicole.sator,|
\urldef{\mailsc}\path|erika.siebert-cole, peter.strasser, lncs}@springer.com|
\newcommand{\keywords}[1]{\par\addvspace\baselineskip
\noindent\keywordname\enspace\ignorespaces#1}

\begin{document}

\mainmatter  

\title{Margin-Regularized Structured Semantic Alignment for Brain-Language Correspondence}
\titlerunning{Margin-Regularized Structured Semantic Alignment}

\author{Jiaqi Wang, Huawen Hu, Shu Zhang}

%
%
%

\institute{School of Computer Science and Technology, Northwestern Polytechnical University, Xi'an 710072, China}

%
%

\toctitle{Lecture Notes in Computer Science}
\maketitle

\begin{abstract}
With the rapid advancement of large language models, brain-language decoding has achieved remarkable progress. However, it remains unclear whether decoded content genuinely reflects neural representations or is largely reconstructed by the language model itself. This ambiguity limits interpretability and hinders the investigation of intrinsic brain-language correspondence. To address this challenge, we propose MD-SigLIP. This margin-regularized structured semantic alignment framework directly aligns brain embeddings with text embeddings in a shared semantic space, enabling retrieval-based decoding. This formulation enables explicit modeling of the correspondence between neural representations and language semantics.
Building upon duplicate-aware sigmoid contrastive learning, we introduce a listwise margin-regularized term that enforces structured ranking constraints between positive semantic clusters and negative samples. By modeling multi-positive semantic structure and margin-based ordering simultaneously, the method captures the manifold organization of language embeddings reflected in neural signals. Experiments demonstrate state-of-the-art retrieval performance under both full-vocabulary and subset evaluation settings.



\end{abstract}
\section{Introduction}

Decoding language from non-invasive brain signals has long been a central goal in computational neuroimaging~\cite{zhang2021survey,tang2023semantic,caucheteux2022brains}. With the rapid development of large language models (LLMs), recent studies have demonstrated increasingly fluent brain-language decoding by integrating neural representations with pretrained generative models~\cite{willett2021high,wang2024enhancing,mishra2025thought2text,caucheteux2022deep}. However, despite impressive qualitative outputs, it remains unclear whether the decoded semantic content genuinely originates from neural signals or is largely reconstructed by the statistical prior of the LLM~\cite{bzdok2024data,jo2024eeg,d2025towards}. This ambiguity limits interpretability and weakens neuroscientific conclusions about how language meaning is represented in the brain.

To more directly investigate the intrinsic relationship between brain signals and language representations, recent work has reformulated decoding as a retrieval-based alignment problem~\cite{d2025towards,liu2025beyond,rashad2023rbqe}. Instead of relying on generative models, these approaches learn a shared embedding space and retrieve text candidates based on similarity to neural representations, enabling explicit measurement of brain-text correspondence.

In previous works, contrastive learning has provided a natural framework for multimodal alignment~\cite{ma2025brainclip,feng2023aligning,ren2021cogalign,hafner2021clip}. The CLIP~\cite{radford2021learning} objective aligns paired representations using a softmax-based contrastive loss, successfully bridging image and text modalities. However, CLIP enforces strict one-to-one matching and introduces strong competition among candidates via softmax normalization. 
Such assumptions may be suboptimal for neural decoding, where semantic representations are continuous and potentially overlapping. SigLIP~\cite{zhai2023sigmoid} replaces softmax with independent sigmoid-based binary cross-entropy, enabling multi-label supervision and improved training stability. While this relaxation mitigates excessive competition, it still treats each pair independently and does not explicitly model relative semantic ordering. To further account for semantic redundancy, duplicated SigLIP (D-SigLIP)~\cite{d2025towards} introduces multi-positive targets by identifying semantically similar text embeddings. This extension captures local semantic clusters in language space and relaxes strict one-hot supervision. 

Motivated by the need to model structured brain-language correspondence, we propose MD-SigLIP (Margin-Regularized D-SigLIP). MD-SigLIP extends D-SigLIP by introducing a listwise margin-regularized term that enforces structured ordering in the embedding space. Specifically, we aggregate multiple semantically similar positives into a cluster-level reference score and explicitly penalize any negative sample that ranks above this positive cluster. This margin-based regularization transforms independent pairwise alignment into a structured semantic ordering problem, encouraging brain embeddings to respect the manifold organization of language representations. As a result, MD-SigLIP simultaneously models multi-positive semantic clusters and enforces structured ranking consistency within each retrieval list.
Experiments on two MEG-language datasets demonstrate that MD-SigLIP achieves superior performance without relying on generative language models. The main contributions of this work are summarized as follows:

\begin{enumerate}
\item We propose a novel alignment framework MD-SigLIP that extends D-SigLIP with listwise margin regularization, enabling cluster-aware semantic modeling and explicit ranking constraints within the embedding space.
\item We provide empirical evidence of semantic and structural consistency between brain and text representations through retrieval-based decoding and representational similarity analysis.
\item We demonstrate that non-invasive neural signals encode retrievable language-related semantic structures, offering an interpretable framework for studying brain-language correspondence.
\end{enumerate}

\section{Method}

\subsection{Problem Formulation}

\begin{figure*}[htbp]
\centering
\includegraphics[width=1\textwidth]{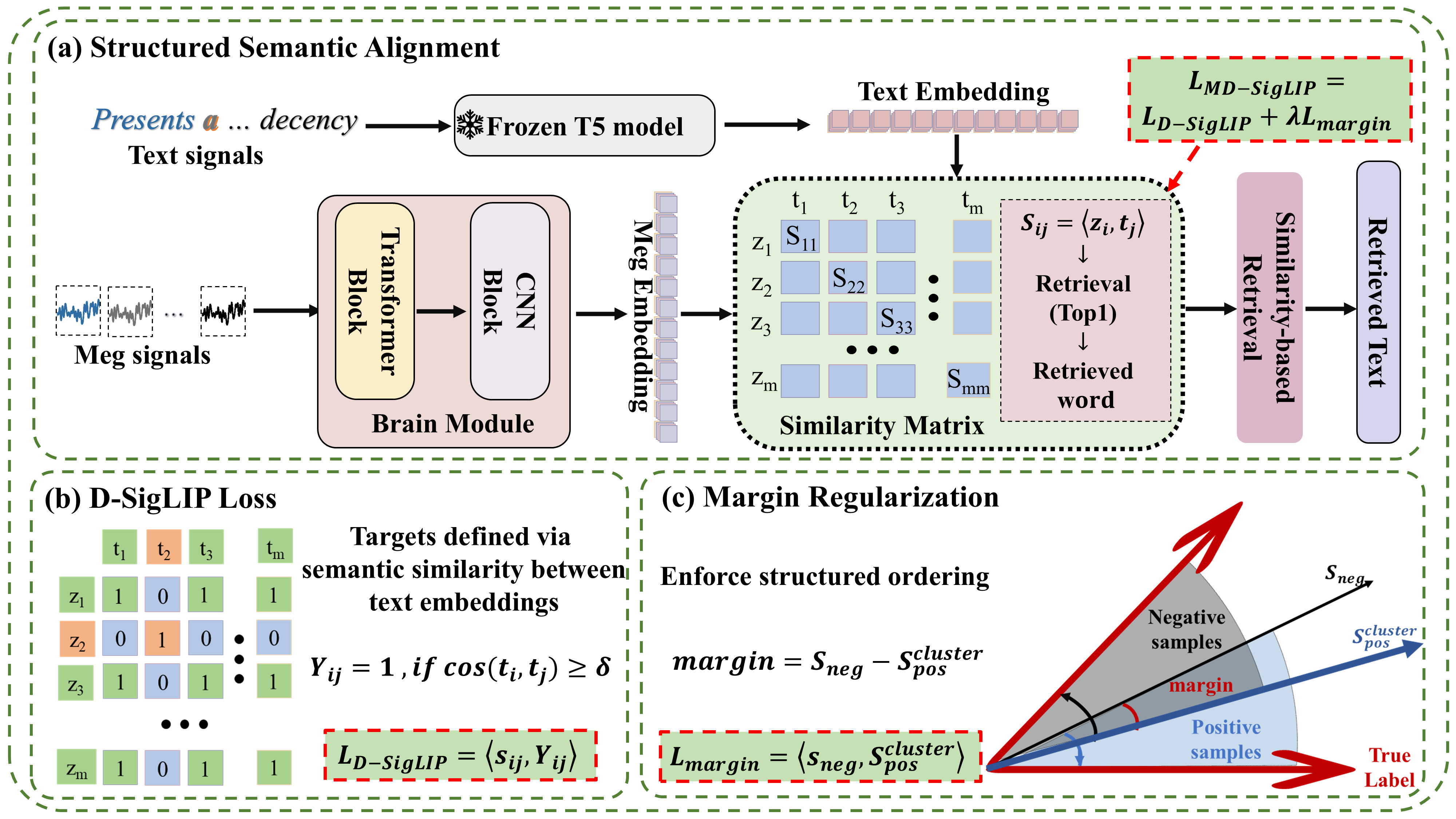}
\caption{Overview of MD-SigLIP.}
\label{fig1}
\end{figure*}

Fig.~\ref{fig1} illustrates the proposed MD-SigLIP framework. 
We reformulate brain-to-text decoding as structured semantic alignment in a shared embedding space. 
Given paired data $\mathcal{D}=\{(x_i,y_i)\}_{i=1}^N$, where $x_i\in\mathbf{R}^{T\times C}$ denotes MEG signals and $y_i$ the corresponding sentence, a frozen T5 encoder $g(\cdot)$ produces text embeddings $t_i=g(y_i)\in\mathbf{R}^d$, while a brain encoder $f_\theta$ maps neural signals into the same space, $z_i=f_\theta(x_i)$. 
Decoding is performed via retrieval based on similarity between $z_i$ and candidate embeddings $t_j$.

To model semantic redundancy, we adopt duplicate-aware supervision (Fig.~\ref{fig1}(b), treating semantically similar text embeddings as multi-positive targets. To further enforce structured ordering, we introduce a listwise margin-regularized term (Fig.~\ref{fig1}(c), encouraging positive clusters to rank above negatives.





\subsection{D-SigLip Loss}

As shown in Fig.~\ref{fig1}(b), followed by D-SigLIP~\cite{d2025towards}, similarity scores are defined as
\begin{equation}
s_{ij} = \tau \langle \tilde{z}_i, \tilde{t}_j \rangle + b,
\end{equation}
where $\tilde{z}_i$ and $\tilde{t}_j$ optionally denote $\ell_2$-normalized embeddings, 
$\tau$ is a learnable temperature, and $b$ a bias term.
Let $S \in \mathbf{R}^{B \times B'}$ denote the similarity matrix within a mini-batch.


Standard contrastive learning assumes one-to-one matching. 
However, semantic representations form continuous clusters. 
We therefore define semantic duplicates in the text space.

Pairwise cosine similarity:
\begin{equation}
C_{jk} = \frac{\langle t_j, t_k \rangle} {\|t_j\| \|t_k\|}.
\end{equation}

Given threshold $\delta$, the duplicate mask is
\begin{equation}
M_{jk} = \mathbf{1}(C_{jk} \ge \delta).
\end{equation}

For each sample $i$, the positive set becomes

\begin{equation}
\mathcal{P}_i = \{ j \mid M_{ij} = 1 \}.
\end{equation}


Following sigmoid-based contrastive learning, we adopt a multi-label binary cross-entropy objective~\cite{d2025towards}:
\begin{equation}
L_{\text{D-SigLip}} = \frac{1}{B} \sum_{i=1}^{B} \sum_{j=1}^{B'} \ell_{\text{BCE}}(s_{ij}, Y_{ij}),
\end{equation}
where $Y_{ij}=M_{ij}$ and
\begin{equation}
\ell_{\text{BCE}}(s,y) = - y \log \sigma(s) - (1-y)\log(1-\sigma(s)).
\end{equation}

This formulation removes softmax competition and supports multiple positives per sample.
\subsection{MD-SigLIP Loss}

While $L_{\text{D-SigLip}}$ models pairwise alignment, it does not explicitly enforce ordering. 
We therefore aggregate positive similarities as 
$s_i^{\text{pos}}=\log\left(\frac{1}{|\mathcal P_i|}\sum_{j\in\mathcal P_i}e^{s_{ij}}\right)$ 
and define negatives $\mathcal N_i=\{j\mid Y_{ij}=0\}$.

The margin regularization term is
\begin{equation}
L_{\text{margin}}=\frac{1}{B}\sum_{i=1}^{B}
\mathrm{softplus}\!\left(\log\!\sum_{j\in\mathcal N_i}
e^{\,s_{ij}-s_i^{\text{pos}}}\right),
\end{equation}
which approximates $\max_{j\in\mathcal N_i}(s_{ij}-s_i^{\text{pos}})$ in a differentiable form and enforces structured ranking.

The final objective is
\begin{equation}
L_{\text{MD-SigLip}}=
L_{\text{D-SigLip}}+\lambda L_{\text{margin}},
\end{equation}
where $\lambda$ balances alignment and margin regularization.

\section{Experiments and Results}

\subsection{Datasets and Settings}

We evaluate our method on two publicly available MEG-Language datasets covering auditory and reading paradigms. Armeni2022~\cite{armeni202210} is an MEG-Audio dataset including 3 subjects with approximately 10 hours of recording each, comprising 7,282 distinct words. SchoffelenRead2019~\cite{schoffelen2019mother} is an MEG-Reading dataset containing 99 subjects and 2,223 distinct words. MEG signals were filtered (0.1--40 Hz), downsampled to 50 Hz, baseline-corrected, and segmented into 3-second word-level epochs. Text representations were extracted using a frozen T5-large encoder. Models were trained for up to 50 epochs using Adam (learning rate $1\times10^{-4}$) with batch size 128 on a single RTX 4090 GPU. Performance was evaluated under a retrieval setting using Top 1 accuracy, Top 10 accuracy, and median rank, computed on both the 250 most frequent words and the full vocabulary.


\subsection{Comparative Analysis of MEG-to-Text Performance across Datasets}

As shown in Table~\ref{tab:combined} (SchoffelenRead2019), our method achieves the best Top 10 accuracy and median rank. 
On the Top 250 setting, Top 10 accuracy improves from 8.2\% (D-SigLIP) to 10.3\%, and median rank decreases from 86 to 65. Under the full vocabulary setting, Top 10 accuracy increases from 6.3\% to 8.7\%, with median rank improving from 180 to 138. Our method slightly but consistently outperforms existing approaches on the Armeni2022 dataset. This is likely because a paradigm with fewer subjects but extensive recordings per subject yields a higher signal-to-noise ratio. While this allows all baseline methods to perform relatively well, our method still establishes a new state-of-the-art.

\begin{table*}[htbp]
\centering
\caption{Comparison of methods on SchoffelenRead2019 and Armeni2022.}
\label{tab:combined}
\setlength{\tabcolsep}{4pt}
\resizebox{\textwidth}{!}{
\begin{tabular}{l ccc ccc ccc ccc}
\toprule
\multirow{3}{*}{Methods} 
& \multicolumn{6}{c}{SchoffelenRead2019} 
& \multicolumn{6}{c}{Armeni2022} \\
\cmidrule(lr){2-7} \cmidrule(lr){8-13}
& \multicolumn{3}{c}{Top 250} 
& \multicolumn{3}{c}{Full} 
& \multicolumn{3}{c}{Top 250} 
& \multicolumn{3}{c}{Full} \\
\cmidrule(lr){2-4} \cmidrule(lr){5-7}
\cmidrule(lr){8-10} \cmidrule(lr){11-13}
& Top1$\uparrow$  & Top10$\uparrow$  & Rank$\downarrow$ 
& Top1$\uparrow$  & Top10$\uparrow$  & Rank$\downarrow$  
& Top1$\uparrow$  & Top10$\uparrow$  & Rank$\downarrow$ 
& Top1$\uparrow$  & Top10$\uparrow$  & Rank$\downarrow$  \\
\midrule
CLIP~\cite{radford2021learning} 
& \textbf{4.4}\% & 8.2\% & 140 
& \textbf{3.2}\% & 5.6\% & 267
& 3.7\% & 17.5\% & 84
& 2.4\% & 9.5\% & 76.5 \\

SigLIP~\cite{zhai2023sigmoid} 
& 1.1\% & 7.7\% & 94
& 0.8\% & 5.9\% & 194
& 15.4\% & 47.3\% & 11
& 10.1\% & 30.4\% & 63 \\

D-SigLIP~\cite{d2025towards}
& 1.3\% & 8.2\% & 86
& 0.8\% & 6.3\% & 180
& 17.2\% & 48.0\% & 11
& 11.4\% & 31.0\% & 62 \\

Ours 
& 2.3\% & \textbf{10.3}\% & \textbf{65}
& 1.2\% & \textbf{8.7}\% & \textbf{138}
& \textbf{17.3}\% & \textbf{48.9}\% & \textbf{10}
& \textbf{11.5}\% & \textbf{31.6}\% & \textbf{56} \\
\bottomrule
\end{tabular}
}
\end{table*}

\subsection{Retrieval Analysis of MEG-to-Text Decoding}
\begin{figure*}[htbp]
\centering
\includegraphics[width=1\textwidth]{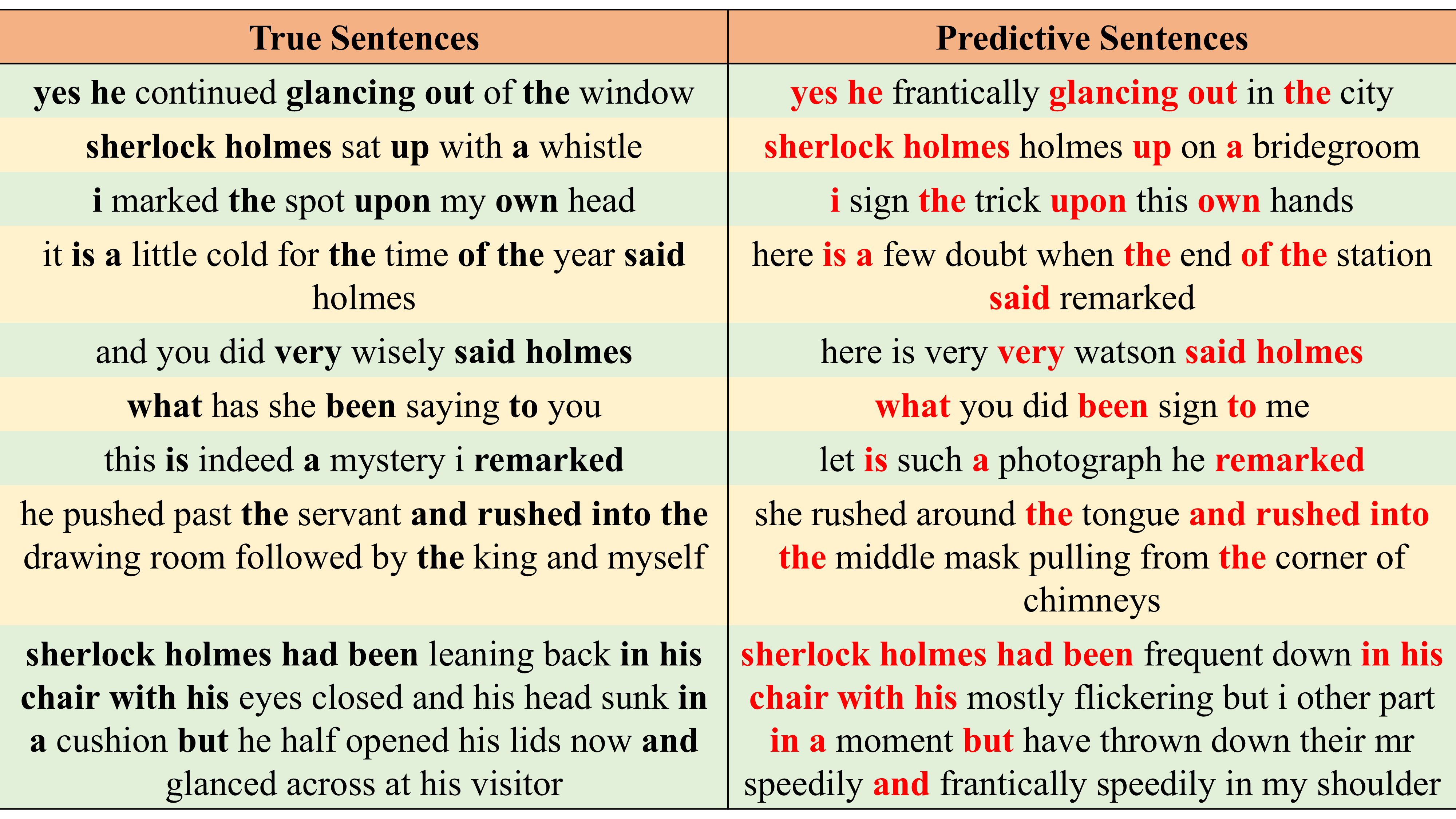}
\caption{Example sentence-level retrieval results. Ground-truth words are shown in bold black, while the corresponding predicted words are highlighted in bold red.}
\label{fig2}
\end{figure*}

To intuitively illustrate the practical decoding capabilities of our model, we provide qualitative examples of sentence-level decoding in Fig.~\ref{fig2}. The results demonstrate that, even in the highly challenging task of decoding continuous MEG signals, our model successfully captures the core semantics and reconstructs syntactically coherent sentence structures. Notably, even when exact word matches fail, the predicted words frequently exhibit high semantic equivalence and syntactic consistency with the ground-truth words (e.g., substituting synonymous nouns or verbs). This indicates that the model is not merely memorizing the training data, but rather learning the high-level semantic representations underlying brain activity.

\begin{figure*}[htbp]
\centering
\includegraphics[width=1\textwidth]{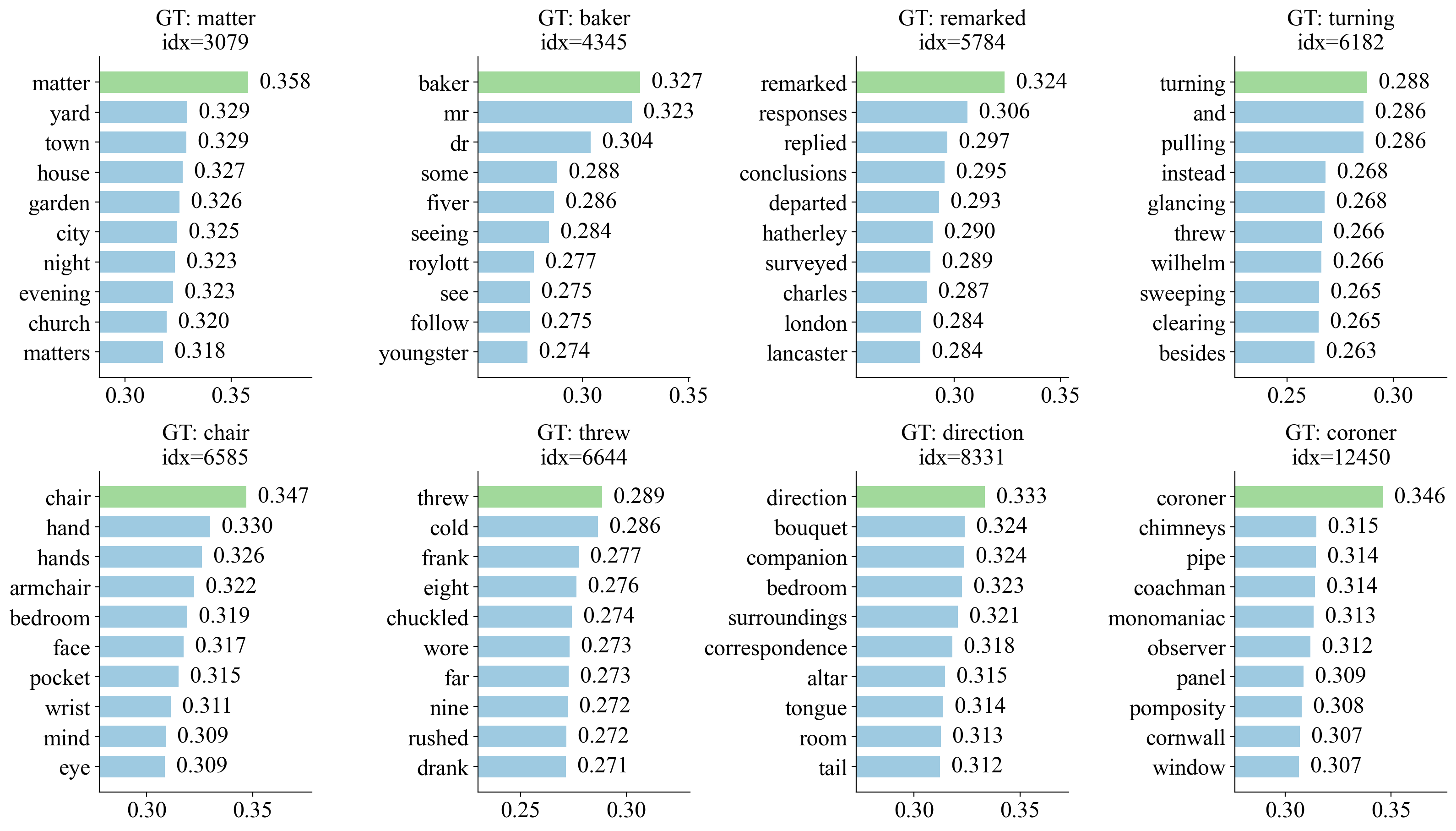}
\caption{Top 10 retrieved candidate words ranked by similarity.}
\label{fig3}
\end{figure*}

Furthermore, Fig.~\ref{fig3} visualizes the Top 10 candidate words retrieved by specific MEG signal segments based on similarity scores. It can be clearly observed that these high-scoring candidates exhibit a significant clustering effect within the semantic space (e.g., vocabulary clusters centered around specific contexts or concepts). This observation suggests that the proposed method learns a smooth and well-structured cross-modal latent space. Within this space, similar brain activation patterns are accurately mapped to neighboring linguistic representations, thereby endowing the model with superior retrieval robustness.

Overall, the retrieval examples and Top 10 candidate analyses consistently demonstrate that the proposed model captures coherent semantic neighborhoods in the shared embedding space. 
Rather than producing isolated word matches, the model maps similar neural activation patterns to semantically related linguistic representations, indicating robust semantic alignment between brain signals and text.

\subsection{Representational Similarity Analysis between MEG and Text Embeddings}

\begin{figure*}[htbp]
\centering
\includegraphics[width=1\textwidth]{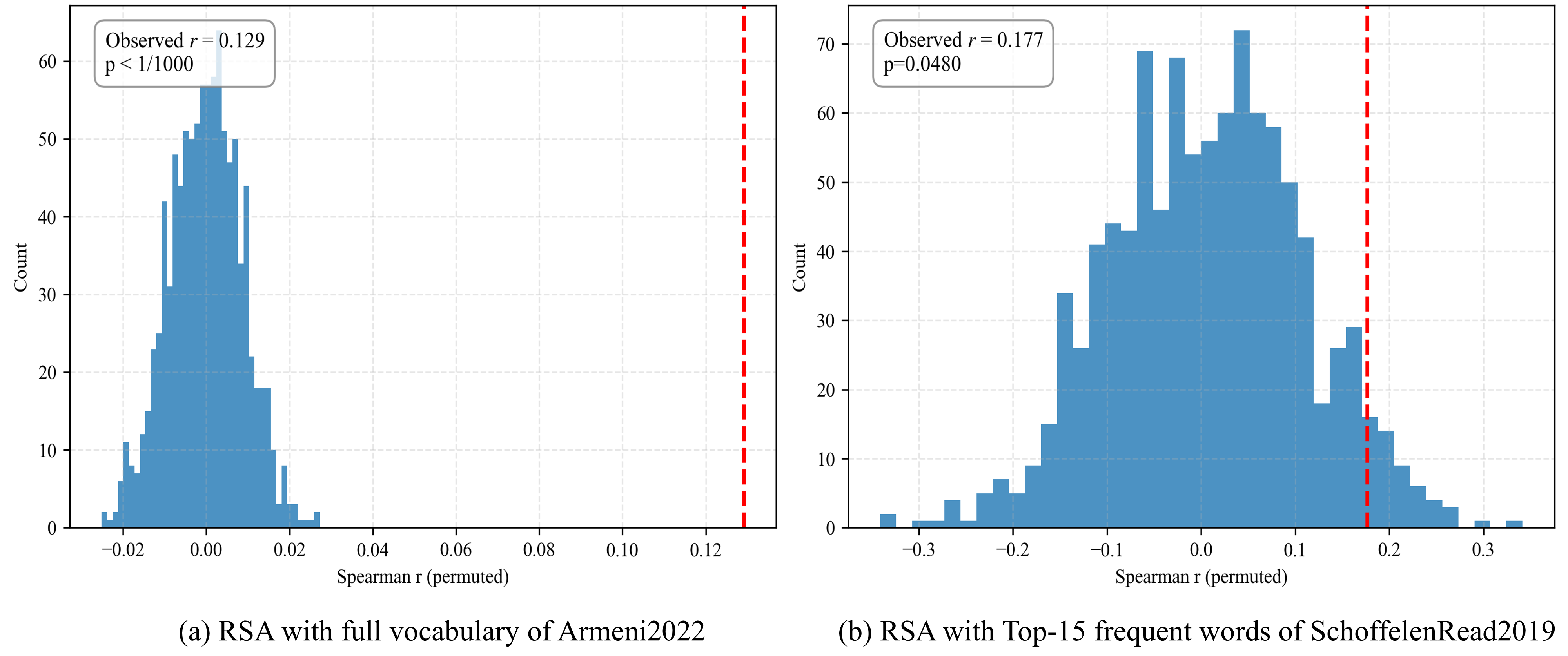}
\caption{Permutation baseline distributions for RSA. Red dashed lines indicate the observed Spearman correlation.}
\label{fig4}
\end{figure*}

To investigate the structural correspondence between MEG neural recordings and text embeddings, we perform representational similarity analysis (RSA)~\cite{kriegeskorte2008representational}. 
Specifically, we compute pairwise similarity matrices within the brain embedding space and the text embedding space, and measure their Spearman correlation. 
To assess statistical significance, we construct permutation-based null distributions by randomly shuffling the correspondence between samples (Fig.~\ref{fig4}).

The permutation distributions serve as statistical baselines, quantifying the range of correlations expected under random alignment. 
A significant deviation of the observed correlation from this null distribution indicates that the similarity structure in the brain embedding space is unlikely to arise by chance.

As illustrated in Fig.~\ref{fig4}(a), the RSA evaluated on the full vocabulary of Armeni2022 yields a highly significant positive correlation ($p < 0.001$), with a tightly concentrated null distribution due to the large sample size. 
This suggests that the geometric structure of brain embeddings closely reflects that of language representations.

Fig.~\ref{fig4}(b) presents the RSA results for the Top 15 most frequent words in SchoffelenRead2019. 
Although the null distribution is broader due to the limited sample size ($T=15$), the observed correlation remains statistically significant ($p < 0.05$). 
This indicates that even within high-frequency vocabularies, stable structural correspondence between neural and linguistic representations can be reliably captured.

Together, these findings provide evidence that MEG signals preserve relational semantic structure that is statistically aligned with language embeddings. 
Moreover, experimental paradigms with extensive recordings per subject appear particularly beneficial for revealing such cross-modal structural alignment.

\section{Conclusion}

In this work, we reformulated brain-to-text decoding as a structured semantic alignment problem. To address the ambiguity introduced by LLM-based decoding, we proposed MD-SigLIP, a margin-regularized structured semantic alignment framework that enforces cluster-aware and order-aware constraints in a shared embedding space. By combining multi-positive supervision with listwise margin regularization, our method captures structured brain-language correspondence and enables interpretable retrieval-based decoding. Experiments on two MEG-language datasets demonstrate state-of-the-art retrieval performance, while qualitative examples and RSA analysis provide converging evidence of semantic and structural alignment between neural and language representations. These findings suggest that neural signals encode retrievable and structured semantic information, offering a framework for studying brain-language correspondence.



%
%
%
\bibliographystyle{splncs}
\bibliography{mybib}

\end{document}